\documentclass[preprint,12pt]{elsarticle}

\usepackage{amssymb,amsthm}
\usepackage{url} 
\usepackage{amsmath}

\usepackage{enumerate}

\begin{document}

\begin{frontmatter}

\title{Topologically sensitive metaheuristics
}

\author{Aleksandar Kartelj \fnref{matf}}
\ead{kartelj@matf.bg.ac.rs}

\author{Vladimir Filipovi\' c \fnref{matf}}
\ead{vladaf@matf.bg.ac.rs}

\author{Sini\v sa Vre\' cica \fnref{matf}}
\ead{vrecica@matf.bg.ac.rs}

\author{Rade \v Zivaljevi\' c \fnref{mi}}
\ead{rade@mi.sanu.ac.rs}

\address[matf]{Faculty of Mathematics, University of Belgrade, Studentski trg 16/IV, 11 000 Belgrade, Serbia} 
\address[mi]{Mathematical Institute, Serbian Academy of Sciences and Arts, Kneza Mihaila 36/III, 11 000 Belgrade, Serbia}
 

\begin{abstract}
This paper proposes topologically sensitive metaheuristics, and describes conceptual design of topologically sensitive Variable Neighborhood Search method (TVNS) and topologically sensitive Electromagnetism Metaheuristic (TEM).
\end{abstract}


\end{frontmatter}


\section{Introduction}

The main purpose of this paper is to address the possibility of using topological theory in the design of metaheuristics. We present the conceptual design of two {\it topologically sensitive metaheuristics}: 

\begin{enumerate}
  \item Topologically Sensitive Variable neighborhood search (TVNS) and
  \item Topologically Sensitive Electromagnetism metaheuristics (TEM).
\end{enumerate}

Our intention is to show that this topological {\it enhancement} can be done in general case, therefore, we select two complementary techniques: VNS is single-solution based and discrete coded metaheuristic, while EM population-based and real coded metaheuristic. The usability of such metaheuristics and their theoretical aspects will be discussed in further papers.

The rest of the paper will cover aspects of scientific disciplines relevant for the proposed idea. Firstly, we will discuss very briefly about elements of Algebraic topology. After that, we will introduce the conceptual design of both TVNS and TEM. 

\section{Computational algebraic topology}

{\bf Simplicial complexes} are combinatorial objects (abstract schemes) used from the early days of algebraic (combinatorial) topology as a bookkeeping device for triangulations of geometric objects. Conversely, they are used in the opposite direction for geometric presentation (visualization, geometric analysis and quantification) of the information (databases, point clouds) of any kind (not necessarily of geometric origin). The importance and versatility of simplicial complexes is illustrated by the fact that they appear under different names and in disguise in different areas of science, in and outside of mathematics. In cooperative game theory they are known as "simple games" (after John von Neumann and Oskar Morgenstern \cite{vonneumann47}). A similar use is in social choice theory (reliability theory). We meet them as threshold complexes (of "short sets"), both in weighted voting games and in the geometry of configuration spaces of polygonal linkages (protein chains). Closely related concepts are monotone hypergraphs, monotone Boolean functions, finite partially ordered sets, etc.

For illustration, the hemi-icosahedron (Figure 1), triangulates the real projective plane and can be used for a combinatorial analysis of this object (homology calculation, non-embeddability in the 3-space, etc.). On the other hand, it provides an important example of a cooperative voting scheme (simple game) for six persons (parties), with 10 winning and 10 losing, 3-element coalitions, which is not realizable as a weighted voting scheme.

\begin{figure}[h!]
\centering
\includegraphics[scale=1]{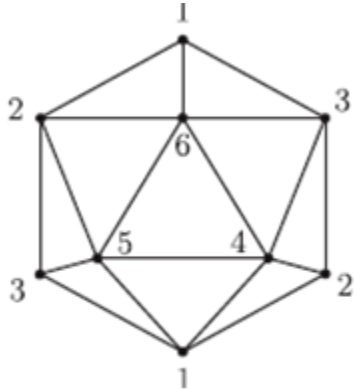}
\caption{hemi-isosahedron}
\label{hemi-isosahedron}
\end{figure}

Simplicial complexes provide historically the first foundation for the theory of “homology groups”, which capture the idea of higher (dis)connectivity (voids, holes) in geometric object. For example, the edge path 1--6--4--1  surrounds an essential 1-hole in the hemi-icosahedron, while if we traverse this edge-path twice, and perturb it to the edge-path 1—-6—-4—-2—-5—-1 (in the same homology class), we obtain a trivial $1$-cycle (illustrating the torsion phenomenon in homology groups).

The so called “persistent homology” (initiated by A. Zamorodian, H. Edelsbrunner, G. Carlsson, and others) opened a new chapter of applications of homological methods and created a large part of what is today known as “applied and computational algebraic topology”. The importance of persistent homology is in its applicability to “dynamical simplicial complexes” (complexes depending on a parameter) which arise in the analysis of large data bases (large finite metric spaces, point clouds, sparse matrices) \cite{zomorodian05}. Information (databases) collected from biological systems is typically less structured, and can be organized, via a concept of “clustering” \cite{carlsson09} into a dynamical (filtered) simplicial complex. 

Theoretical framework for “clustering”, i.e. for creating (parameterized) simplicial complexes out of point cloud (a finite metric subspace of Euclidean space) is provided by the concept of the “nerve of a covering”, alpha-shapes, Vietoris-Rips complexes, etc. Algebraic counterpart of this construction is the concept of a $Z[t]$ (differential) module $M$ (rather than $Z$ differential module in classical homology), where the operator $t$ corresponds to the dynamical part (filtration) of the simplicial complex. The operator $t$ commutes with the boundary operator on $M$ (from standard simplicial homology) so the structure of a $Z[t]$ module descends to the (graded) homology group $H(M)$. This allows us to discriminate {\bf long-lived} elements (cycles) in $H(M)$ from short lived. Each cycle $x$ in $M$ is born and after several iterations of the operator $t$ it disappears. If its lifespan (number of $t$ –- iterations before it is annihilated) is very small, the cycle $x$ is interpreted as the “noise” i.e. it does not provide any significant information about the structure of the original database (point cloud). In the opposite case, the cycle is relatively stable (long lived) under $t$ – iterations and provides insight into some important structural feature of the original point cloud. All long-lived cycles together with the intervals corresponding to their life spans define the so-called {\bf barcode} which summarizes the persistent homology information about the original point cloud.

Figure 2 (taken from \cite{ghrist14}) exemplifies a “point cloud” shaped as a surface with two visible holes (homology cycles). 

\begin{figure}[h!]
\centering
\includegraphics[scale=1]{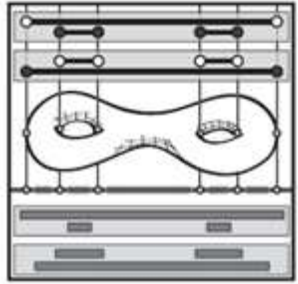}
 \caption{point-cloud}
\label{point-cloud}
\end{figure}

If the cloud (surface) is scanned from left to right (for the level set persistence), there emerge significant intervals of persistence creating the associated “barcode” of the point cloud. Finding intervals of persistence out of a point cloud can be compared to the “peak detection” in time-series analysis, or to finding of significant Fourier coefficients of a function. In all these cases we expect to find information about essential features of the analyzed object (point cloud, time series, function, etc.).

\section{Metaheuristics and topology}

The main motivation for integrating topology and metaheuristics comes from the notion that metaheuristics might use the topological regularities inside the solution space to {\it better maneuver} through it. This can become especially useful when the solution space becomes extremely large. In such situation classical metaheuristics might use too much resources in order to search the solution space. Although this sounds like it could lead to premature convergence to local optima, we stress that our conceptual design essentially generalizes and encompasses the classical metaheuristic algorithms. This means that the proposed metaheuristics, during its execution, gradually converge to its classical variants. Also, by imposing adequate parameters, these topologically sensitive metaheuristics can be used as a classical through its whole execution.

Execution of proposed topologically sensitive metaheuristics will resemble execution of any other metaheuristic: each execution creates a path in fitness landscape in order to reach global optima and avoids local optima. Therefore, fitness landscape analysis, which includes analysis of local optima positions, is very important for design of such metaheuristics. In other words, if some topological regularity in fitness landscape is detected, that regularity can be exploited and used for designing metaheuristic that will perform better than the alternatives. Topology-based models and techniques already achieved good results in revealing hidden structures and detecting new regularities \cite{Reimann17} and \cite{Lee17}, so it can be expected that it will be helpful in this domain.        

We address design and implementation considerations of two topologically sensitive metaheuristics: 
\begin{itemize}
\item TVNS (topologically sensitive Variable neighborhood search) and
 
\item TEM (topologically sensitive Electromagnetism metaheuristic). 
\end{itemize}

\subsection{Conceptual design of TVNS}

Variable neighborhood search (VNS) algorithm is a well-known metaheuristic optimization approach introduced by Mladenovi\'{c} and Hansen \cite{Mladenovic97}. The basic strategy of the VNS is to perform the search inside the neighborhoods of the current best solution. In order to avoid being trapped in local suboptimal solutions, VNS systematically changes the neighborhoods, following the empirical observation that multiple local optima are often in correlation, including the fact that the global optimum is also a local optimum with respect to all neighborhoods. VNS is successfully used for solving various NP-hard problems of great practical importance \cite{Hansen10} (location problems, graph coloring problems \cite{Matic17}, knapsack and packing problems, vehicle routing problems). VNS is also successfully applied to Data mining domain, to design problems in communication and problems in biosciences and chemistry \cite{Grbic19a}. 
 VNS consists of the following steps: 
\begin{itemize}
\item In the initialization part of the algorithm, the control parameters (minimal and maximal VNS neighborhood structure size, finishing criteria) are specified, and initial solution is generated in pseudo-randomly manner. This solution also becomes the current best.
\item After the input and initialization procedures, the algorithm enters the main loop. In each iteration of the main loop, the following procedures are performed:
  \begin{itemize}
	\item	Shaking - in order to escape local suboptimal solutions, a new solution within a parametrized neighborhood of the current best solution is generated. 
  \item Local search - starting from the new solution obtained in the previous step, other possible solutions within local neighborhood are systematically examined with the aim of finding the local optimum.
	\item Neighborhood change - depending on the success of the previous two procedures, the current neighborhood size is adjusted. More precisely, when the current best solution is changed, neighborhood size is reduced to minimal, otherwise it is cyclically increased by 1 (cycle ends at maximal neighborhood size).   
  \end{itemize}
\item	The iteration process ends when the maximal number of iterations is reached, the same best solution is not changed for the predefined maximal number of iterations, etc.
\end{itemize}

The local search step performs search intensification while the shaking step is related to the diversification of search. Moreover, larger neighborhood size directs toward stronger diversification while smaller neighborhood size forces intensification within search space.

When looking at the VNS problem solutions from topological point of view, it can be observed that they can be modeled as $0$-simplices. Therefore, the collection of VNS solutions forms a point cloud. Classical VNS works only on top of $0$-simplices and further uses $1$-simplex neighborhoods for solution transitions. The shaking procedure therefore moves the current (best) solution to some other solution that is edge-connected with respect to given distance function. 

For example, if we have two binary-coded solutions $1011110$ and $1111110$ and Hamming distance function, then these two $0$-simplices are connected in $1$-simplex when their Hamming distance is $1$. Therefore, the movement (during the shaking) from the first one to the second one is possible. Note that distance function is parametrized with $k$ (neighborhood size) inside classical VNS algorithm. For $0$-simplex VNS (classical VNS) we need one current solution in order to generate a sequence of new ones during the shaking procedure.
 
\textbf{Notation remark}: TVNS is essentially conceived as a generalization of VNS that builds on $m$-simplex data (with special case $m=0$ being a classical VNS). We will also sometimes refer to $(m+1)$-simplex neighborhood which corresponds to collection of all valid simplices that can be formed by adding $0$-simplex to observed $m$-simplex. Therefore, $1$-simplex neighborhood correspond to classical VNS, while $m$-simplex neighborhoods where $m>1$ refer to its topological generalizations. 

The main adjustment needs to be made inside shaking procedure where a new solution should be generated based on the set of previous solutions. When, for example, $2$-simplex neighborhood is used with Hamming distance strictly $2$ (or at most $2$) we need at least $2$ previous problem solutions in order to generate the new one. In the case when the distance is strictly $2$ and previous solutions are $1011110$ and $1111111$, we can, for example, generate a new solution $1011011$ by randomly picking it from the set of possible alternatives {$1011101$, $1011011$, $1010111$, ...}. Note that these three solutions now form a $2$-simplex with respect to imposed distance function. Similarly, if Hamming distance should be at most $2$, then, for example, by combining solutions $1011110$ and $1111111$ one could get new solution $1011111$. Here, solutions 1 and 2 are at Hamming distance $2$, but all other pairs of solutions are at distance $1$. Also, $2$-simplex neighborhood could be combined with neighborhoods of higher cardinality. For example, when using strict Hamming distance $4$, from solutions $1011110$ and $0111011$ one could get solution $1100111$.  

It is obvious that TVNS will need an additional memory since new solutions cannot be simply generated based on the previous one when $m>0$. We will have to deal with three questions: 
\begin {enumerate}[(a)]
\item	Which solutions should be kept in memory during optimization?
\item	How to select the target simplex that will be extended with new solution (0-simplex)? 
\item	How to generate new valid solution, i.e. the one that extends selected m-simplex to (m+1)-simplex?
\end{enumerate}
There are at least two answers to question (a). The most memory and computationally intensive approach would be to store all previous solutions. The other approach would be to have a fixed memory and thus remove certain solutions when that fixed memory becomes full. 

Selection of target simplex is partially deterministic in a sense that the set of target simplex candidates is formed by collecting all $m$-simplices (within distance function parameterized by $k$) that contain current TVNS solution. The answer to question (b) that is in line with the original essence of shaking procedure is to perform random selection. Note that when $m=0$, the only candidate $0$-simplex is current solution because it is the only $0$-simplex that contains itself, thus TVNS shaking reduces to classical VNS shaking.  

Finally, regarding question (c), when candidate simplex is selected, new solution can be selected in at least two ways. The first would be to randomly select valid solution with respect to selected simplex. The second would be to randomly select best valid solution, i.e. the one of the solutions that forms an $(m+1)$-simplex and additionally optimizes certain criterion, for example, it achieves the most balanced pairwise distances inside $(m+1)$-simplex (note that all distances are less or equal to given $k$). Both approaches are using randomness like classical VNS, however, the randomness is here controlled in order to allow only valid solutions with respect to selected simplex. Specially, when $m=0$, every $k$-inversion of the current solution is within $1$-simplex neighborhood of size $k$.  

As local search step is concerned, the similar generalization as for the shaking step can be done. The difference with shaking is that local search is systematic. This means that unlike selecting a single simplex from possible simplex candidates, local search checks all simplex candidates for given $m$ and $k$. Also, local search operates locally which means that neighborhood size is small, $k=1$, $k=2$ or at most $k=3$. Note that when using combination $m=0$, $k=1$, the classical $1$-swap local search is performed. Larger neighborhoods $k>1$ are mostly avoided in classical VNS local search procedures due to their increasing computational costs - incrementation of $k$ usually increases complexity by linear factor. TVNS can check neighborhoods of size $k>1$ for $m>0$ more efficiently since it performs exhaustive search only with respect to observed $m$-simplices. 

The main loop of TVNS should be made in such a way that the sequence of neighborhood structures, that are now parametrized by $m$ and $k$, starts with the most restrictive neighborhood and after that proceeds with the sequence of more relaxed ones. Therefore, the neighborhoods will start with smallest neighborhood size $k=k_{min}$ and the most restrictive simplex structure $m=m_{max}$, and further proceed with reduction of $m$ by $1$. When $m$ reaches $0$, it basically means that classical VNS algorithm is to be performed. After that, the $k$ is increased by $1$ and $m$ is reset to $m_{max}$. The full cycle through neighborhoods is done when $k$ reaches $k_{max}$ and $m$ reaches $0$. If, at some moment, current solution is improved, both $k$ and $m$ are reset to its initial values. 

The overall effect that we expect TVNS will have on the search process in comparison to VNS is increased preservation of the same or similar topological regularity through time (if this regularity exists). We also believe that the expansion of already existing simplices, especially large ones, is well motivated. This is because the existence of regular formation of local optima itself is an indicator that more new local optima may be found around that formation.  
Another important observation is that since TVNS falls back to classical VNS, we can expect that TVNS will be generally applicable, i.e. if the topological regularity is low and cannot be exploited, TVNS should work at least as good as classical VNS (though performance might get deteriorated). Similar observation can be made for TEM algorithm as well.

There are many more considerations regarding TVNS, for example - why do we expect the preservation of topological regularity and in what relation are TVNS neighborhood structures with persistent homologies? However, since this document only introduces its novel conceptual design those considerations will be done in the future theoretical analysis and empirical studies.

\subsection{Conceptual design of TEM}

The electromagnetism method (EM) is a metaheuristic algorithm which is introduced by Birbil and Fang in \cite{Birbil03}. EM utilizes an attraction-repulsion mechanism to move sample points towards optimality. Each point (particle) is treated as a solution and a charge is assigned to each particle. Better solutions possess stronger charges and each point has an impact on others through charge. The exact value of the impact is given by Coulomb’s Law. This means that the power of the connection between two points will be proportional to the product of charges and reciprocal to the distance between them. In other words, the points with a higher charge will force the movement of other points in their direction more strongly. Besides that, the best EM point will stay unchanged. The charge of each point relates to the objective function value, which is the subject of optimization.
This populational metaheuristics is successfully applied to various problem domains: set covering problems \cite{NajiAzimi10}, hub location problems, maximum betweenness problems \cite{Filipovic13}, for physical mapping with the end probes in molecular biology, for automatic detection of circular shapes embedded into cluttered and noisy images \cite{Cuevas12}, for feature selection within classification \cite{Filipovic17}, etc.
EM algorithm consists of the following steps:
\begin{itemize}
\item	In the initialization part of the algorithm, the control parameters (maximal number of iterations, the number of solution points, and maximal number of repetitions of the same solution) are specified. 
\item	After the input and initialization procedures, the algorithm enters the main loop. In each iteration of the main loop:
	\begin{itemize}
	\item	for each solution point:
		\begin{itemize}
		\item	The objective function is calculated.
		\item	Local search procedure is executed.
		\end{itemize}
	\item	Charges and forces among solution points are calculated, based on the previously calculated solution points objective function values.
	\item	Movement of solution points within the search space is performed. This movement is directed by 1) forces that are calculated in the previous step and 2) with randomness. 
	\end{itemize}
\item	The iteration process ends when either the maximal number of iterations is reached, or the same best solution is not changed for the predefined maximal number of iterations.
\end{itemize}

In order to maintain the search effectiveness of the algorithm, choosing an appropriate representation of the candidate solution plays a key role. Each solution point is usually represented as a $n$-dimensional vector of real valued coordinates.

When comparing classical VNS to classical EM and having in mind previous conceptual design for TVNS, three main differences should be noted. First, EM is based on the real-coded vectors while VNS uses discrete-coded vectors. Second, EM is population-based metaheuristic, unlike VNS that is a single-solution metaheuristic. Third, instead of shaking within VNS, EM is based on the different solution movement operator that uses calculations of charges and forces to determine simultaneous movement of the whole solution population. 

The first difference means that the distance function for TEM needs to be defined differently. Euclidean distance might be the first alternative that we will experiment with. The second difference is favorable since it makes topological enhancement even more natural than in the case of VNS. Here, instead of using historical solutions, the current population of solutions can be used. Finally, due to all these differences, especially the third one, TEM conceptual design needs to be somewhat different from TVNS. The main difference in TEM, in comparison to classical EM, is in the movement step. For each solution point, within TEM, we try to find new solution position inside the solution space that will form a $m$-simplex with other $m$ solution points from the current population (two variants: with or without that solution considered as a candidate). This should be done in such a way that average or maximal distance among $0$-simplices is minimized. Again, as in TVNS, we start with the most restrictive $m$-simplex, i.e. $m=m_{max}$. As mentioned in the description of the EM algorithm, the movement is controlled partially by forces that affect the solution point and partially by the randomness. In TEM, the movement is also controlled by forces, but now the randomness is restricted with respect to parameter $m$. This means that for $m>0$, the set of possible positions from which the new position is randomly chosen now becomes smaller, i.e. it becomes the subset of the set of possible positions where $m=0$ (classical EM). If, for a given solution point and current simplex size $m$, the movement is not possible, the simplex size is reduced by $1$. Finally, if $m$ reaches $0$, that basically means that algorithm fell back to its classical version where every movement within distance threshold is allowed. Note that this process of reducing the simplex size is done for each solution point separately. Also, note that the distance function threshold for deciding whether two $0$-simplices are connected can be implicitly set by an algorithm to a sufficiently large value which will always allow a movement in the classical fallback EM.

As with TVNS, we stress out that this is a novel conceptual design so further theoretical and empirical analysis is yet to be done. 

\subsection{Other considerations}
Prior to implementation phase, some preliminary analysis have to be concluded:
\begin{itemize}
\item	Testing hypothesis whether problem solutions form topological simplicial complexes with certain regularity in their structure. In order to do that for TVNS, we will use standard VNS and remember all solutions, representative sample of all solutions or all near-best solutions. We will further use obtained results as point cloud data, and build simplicial complexes on top of them. If we detect certain regularity in topological structure, then we will use that regularity to improve the solution search. 

\item	Test persistence of the obtained topological structures through concept of persistence homology. For example, what happens when we gradually increase neighborhood size ($k$) in VNS, and how does this influence established topological structures.  
\end{itemize}

Another aspect of creating topologically sensitive metaheuristics will be the analysis of its usability. According to \cite{Wolpert97}, all algorithms that search for an optimum of a cost function perform the same when averaged over all possible cost functions. So, for any search/optimization algorithm, any elevated performance over one class of problems is exactly paid for in performance over another class - there is no optimization algorithm that works excellent for all classes of instances. Therefore, usability of proposed approach should be analyzed on various problems, including NP-hard well-known optimization problem from the literature (e.g. minimum set cover problem, traveling salesman problem).

\section{Conclusions and further research}

In this paper, topologically sensitive metaheuristics are proposed, and conceptual design of topologically sensitive Variable Neighborhood Search method (TVNS) and topologically sensitive Electromagnetism Metaheuristic (TEM) are elaborated.

We would like to emphasize that approach proposed in this paper is inherently general and can be successfully used in many different innovative ways. Due to the generality of the idea of topological sensitivity, there are various levels of generality. Therefore, there is an open space for further research. Concerning the next step in our research, we will focus on following:
\begin{itemize}

\item	Implementation of proposed conceptual designs.

\item	Empirical and theoretical evaluation of the proposed methods. 

\item	Testing applicability of the proposed methods for solving problems in various domains like social network analysis, biological networks, etc.  

\item	Design and implementation of new topologically sensitive methods that are based on other metaheuristics. 

\end{itemize}

\bibliographystyle{elsarticle-num} 
\bibliography{topologicaly-sensitive}

\end{document}